\documentclass{svproc}
\usepackage{url}

\usepackage{csquotes}
\usepackage{amsfonts}
\usepackage{amsmath}
\usepackage{mathtools}
\usepackage{graphicx}
\usepackage{subcaption}
\usepackage[export]{adjustbox}
\usepackage{siunitx}
\usepackage{hyperref}
\usepackage[bottom]{footmisc}

\usepackage[colorinlistoftodos]{todonotes}

\newcommand*\diff{\mathop{}\!\mathrm{d}}

\begin{document}
\mainmatter              
\title{Telexistence and Teleoperation for \\ Walking Humanoid Robots}
\titlerunning{Teleoperation for Humanoid Robots}  
%
\author{Mohamed Elobaid\inst{1} \and Yue Hu\inst{1} \and Giulio Romualdi\inst{1} \inst{2} \and Stefano Dafarra\inst{1} \inst{2} \and \\ Jan Babic\inst{3} \and Daniele Pucci\inst{1}}

\authorrunning{Mohamed Elobaid et al.} 

\institute{Fondazione Istituto Italiano di Tecnologia, 16163 Genova, Italy\\
\email{name.surname@iit.it}
\and
DIBRIS, Universit\`a degli studi di Genova, 16163 Genova, Italy
\and
Jo\~zef Stefan Institute, Jamova cesta 39, 1000 Ljubljana, Slovenia\\
\email{jan.babic@ijs.si}
}

\maketitle              

\begin{abstract}
This paper proposes an architecture for achieving telexistence and teleoperation of humanoid robots.
The architecture combines several technological set-ups, methodologies, locomotion and manipulation algorithms in a novel manner, thus building upon and extending works available in literature. The approach allows a human operator to command and telexist with the robot. Therefore, in this work we treat aspects pertaining not only to the proposed architecture structure and implementation, but also the human operator experience in terms of ability to adapt to the robot and to the architecture. The proprioception aspects and embodiment of the robot are studied through specific experimental results, which are treated in a high-level manner. Application of the proposed architecture and experiments incorporating user training and experience are addressed using an illustrative bipedal humanoid robot, namely the iCub robot.
\end{abstract}

\section{Introduction}
With the advancements in the field of robotics, teleoperation is no longer seen just as a mean of manipulating a device remotely, but also as a possible mean for \textit{telexistence}, i.e. giving a real-time sensation to a human being to be in another place, strictly speaking in the context we are considering, a place in the real world.

Humanoid robots have been popular in research for decades, and have seen several improvements in recent years. With these robots, which are often morphologically and visually similar to humans, teleoperation might be automatically associated with telexistence, and it is somehow expected that the robot would behave in a human-like way. This paper contributes towards the development of a control architecture achieving telexistence of a human operator by means of a humanoid robot.

In the context of telexistence via humanoid robots, motion retargeting plays an important role. The problem of offline motion retargeting and imitation has been addressed quite successfully in the literature \cite{Safonova2003} \cite{Suleiman2008},  both using a marker-based approach and marker-less one, respectively. The extension to real time motion retargeting gave rise to a different set of challenges, in which time-consuming optimization algorithms proved futile (time is a hard constraint), and ways to speed-up computations by means of using different models as in \cite{montecillopuente:tel-00538681} or using task space variables and eliminating the need for inverse kinematic solutions as in \cite{Demircan2006}, were attempted with various success rates.

The architecture for human motions retargeting and humanoids teleoperation that we detail in this paper in an illustrative manner, is inspired by the work of A. Spada et al \cite{10.1007/978-3-319-89327-3_7}. We address the basic theoretical aspects as well as treating
implementation related issues utilizing the iCub humanoid robot platform \cite{Metta2010} \cite{Natale2017},
which uses YARP \cite{Metta2006} as middleware for communications, as our test bench to
perform given manipulation tasks and walking. 
In addition, another aspect which is addressed by this paper, is the operator ability to interact with the robot, and tools in general, extending works by J. Babic et. al. in \cite{Peternel2013} and \cite{doi:10.1177/1059712311411112} to a set-up where the robot is walking. 

And while we don't employ learning algorithms to allow the robot to mimic the human as in the above mentioned papers, our aim is to study the human ability to adapt while \enquote{embodying} the robot, and to what extend our proposed set-up is immersive.

It is also of note that a similar set-up pertaining to the locomotion aspect was suggested recently, to the best of the author's knowledge, only in  \cite{10.1007/978-3-319-89327-3_7}, however, in the aforementioned work, manipulation tasks and complex end-effector motions retargeting were not included, and were instead recommended as possible extension. Furthermore, our set-up allows analysing the full immersion of the user in terms of telexistence via teleoperation as mentioned above.

This manuscript is organized as follows; an overview of the related works and general hints on the structure of the paper were presented in this section. The second section details the proposed retargeting and teleoperation architecture, highlighting the various components and tools used, its novelty and significance. The third section will present aspects of proprioception, i.e. sense of relative position of body segments in relation to other body segments, and human adaptability when teleoperating the robot. The fourth section contains the experiments carried out to illustrate our results, as well as a thorough discussion
highlighting the observed limitations and conveying possible future directions. Some concluding remarks end the paper.
Some aspects hinted at in this manuscript (e.g. walking controllers \cite{8625025}, inverse kinematics and task hierarchies \cite{Nori2015}) will not be detailed for the sake of compactness, however,  remarks, footnotes and references will make this paper as self-contained as possible.

%

\section{Teleoperation Architecture}
\label{sec:teleoperation_architecture}
\begin{figure}[t]
    \centering
    \includegraphics[width=\textwidth]{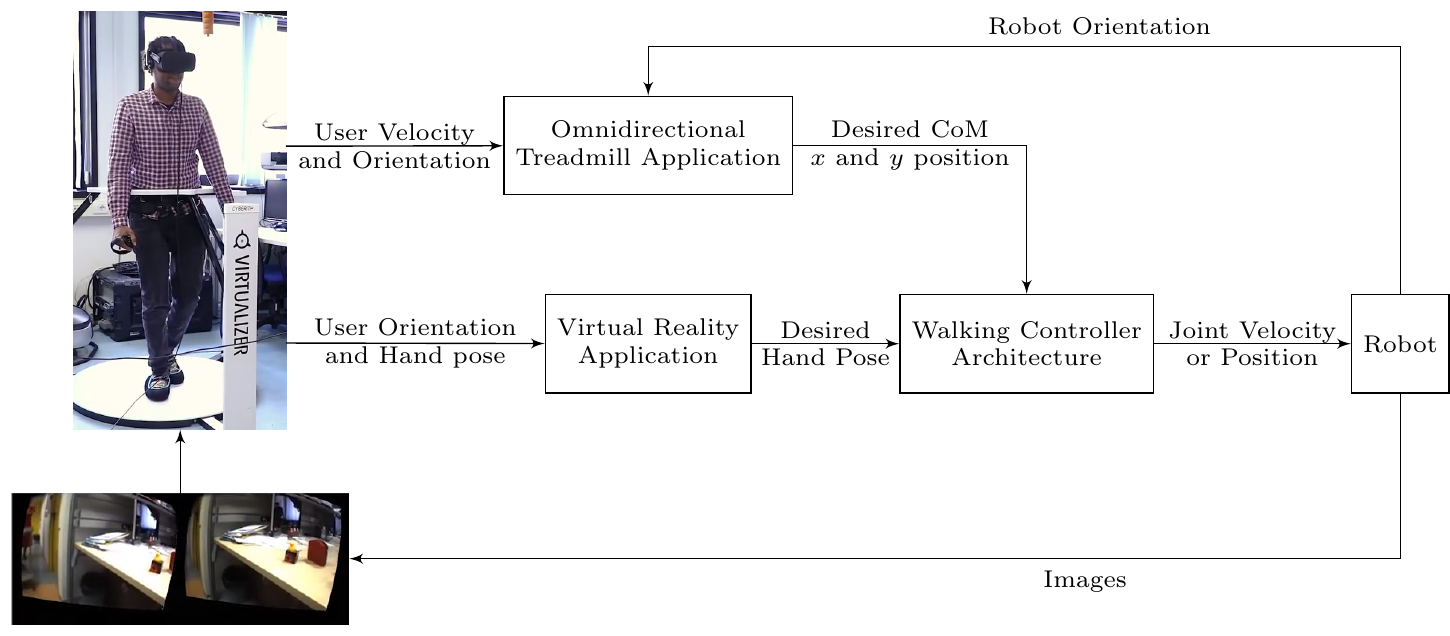}
    \caption{The teleoperation architecture is composed of three blocks: the \emph{Walking Controller Architecture}, the \emph{Omnidirectional Treadmill Application}, and the \emph{Virtual Reality Application}.}
    \label{fig:block-diagram}
\end{figure}

This section introduces the architecture we propose for locomotion and motions retargeting. As depicted in Fig.~\ref{fig:block-diagram}, the user walks in an omnidirectional treadmill wearing a virtual reality system, therefore the architecture consists in the following blocks: the \emph{Walking controller}, \emph{Omnidirectional treadmill application}, and \emph{Virtual Reality application}.  

The outputs of the treadmill\footnote{namely the \emph{Cyberith Virtualizer}, a friction-less platform that allows \enquote{players} to command locomotion of avatars by providing sensors for speed, direction and height \cite{Cakmak2014}} are the walking velocity and the orientation of the user, while the readouts of the virtual reality system\footnote{namely an \emph{Oculus Virtual Reality} set} are the position and the orientation (pose) of the user's hands expressed w.r.t the virtual reality inertial frame, as well as the user's head orientation. 
The purpose of the \emph{virtual reality application} is to manipulate the virtual reality system readouts along with the user orientation and to obtain the desired hand pose. The \emph{Omnidirectional treadmill application} uses the user velocity and orientation along with the orientation of the robot to evaluate the desired position of the robot center of mass (CoM). In details, the desired planar CoM position is expressed in a frame placed in the middle of the feet whose $x$ axis points forward and the $y$ axis points laterally to the left.
The \emph{walking controller architecture} is in charge of guaranteeing the tracking of the desired hands pose and walking direction. 
Finally the images coming from the robot cameras are sent to the virtual reality headset giving to the user the possibility to see what the robot is seeing. 


\subsection{Omnidiretional Treadmill Application}
The Omnidirectional treadmill application evaluates the desired CoM planar position of the robot by using the treadmill readouts and the orientation of the robot.
In details, the desired CoM position is heuristically evaluated by comparing the robot yaw angle $\theta_r$ to the user orientation $\theta_u$, and finally scaled by using the user velocity $v_u$. In equations:
\begin{equation}
    \begin{cases}
    x = v_u \cos(\theta_u - \theta_r) \\
    y = v_u \sin(\theta_u - \theta_r),
    \end{cases}
\end{equation}
where the $x$ represents the frontal direction, and $y$ is pointing left.
This choice of $x$ and $y$ allows the robot to follow the user walking direction, indeed if the two angles are the same the robot will move forward otherwise it will turn.

\subsection{Virtual Reality Application}
The virtual reality application combines the signal of the user's hands pose with the orientation of the user and it sends the data to the walking controller for the retargeting. The hand poses retrieved from the Oculus are expressed with respect to the virtual reality inertial frame $\mathcal{V}$. The center of the frame $\mathcal{V}$ is in general given by the initial position of the headset, while the $z$ axis points upward and the $x$ axis points forward. To perform the hands retargeting one wants to map the desired hand pose into a transformation between the frame placed in the robot hands and robot head. As a consequence, we decided to express the hands pose with respect to a frame rigidly attached to the omnidirectional treadmill and so to the user. This frame, called retargeting frame $\mathcal{R}$, has the same center of the virtual reality inertial frame but the $x$ axis always points to the user forward direction. More formally the following rotation matrix maps the transformation between the two frames $\prescript{\mathcal{V}}{}{R}_\mathcal{R} = R_z (\theta _ u)$. As a consequence, the desired hands pose with respect to the retargeting frame easily follows:
\begin{equation}
    \prescript{\mathcal{R}}{}{T}_{\mathcal{H}^u_\#} = 
    \prescript{\mathcal{R}}{}{T}_\mathcal{V}
    \prescript{\mathcal{V}}{}{T}_{\mathcal{H}^u_\#} =
    \begin{bmatrix}
    R_z(-\theta_u) & 0 \\
    0^\top & 1    
    \end{bmatrix}
    \prescript{\mathcal{V}}{}{T}_{\mathcal{H}^u_\#},
\end{equation}
where $\mathcal{H}^u_\#$, with $\# = \{l, r\}$, is the frame attached to the user hand frame. 
\par
In order to map the human hand position onto the robot platform, the kinematics scaling may be necessary. Here we propose to scale down position components of the transformations generated by the human to that of positions vectors appropriate for the robot by means of scalar multiplication with a suitable user specific ratio (e.g. links lengths ratio between human and robot).
\par
Last but not least, the hands frame and the teleoperation frame have to be mapped onto the robot frames. In details we mapped the teleoperation frame onto a frame placed on the head of the robot $\mathcal{H}$ and the hands frame onto the robot hand frame $\mathcal{H}^r_\#$. Here robot-specific constant homogeneous transformation may be required. 
\par
To summarize, the virtual reality application retrieves the angle $\theta _ u$ from the omnidirectional treadmill and the hands homogeneous transformations $\prescript{\mathcal{V}}{}{T} _ {\mathcal{H}_\# ^ u} $ from the virtual reality system; then it evaluates the $\prescript{\mathcal{H}}{}{T} _ {\mathcal{H}_\#^r}$ transformations.
For the sake of completeness the term of $\prescript{\mathcal{H}}{}{T} _ {\mathcal{H}_\# ^ r}$ is shown:
\begin{equation}
\prescript{\mathcal{H}}{}{T} _ {\mathcal{H}^r_\#} = 
\prescript{\mathcal{H}}{}{T} _ {\mathcal{R}} 
\prescript{\mathcal{R}}{}{T} _ {\mathcal{V}}
\prescript{\mathcal{V}}{}{T} _ {\mathcal{H}^u_\#} 
\prescript{\mathcal{H}^u_\#}{}{T} _ {\mathcal{H}^r_\#}.
\end{equation}

\subsection{Walking Controller Architecture}
The walking controller used for the teleoperation is a three-layer controller architecture \cite{8625025}. 
Here, a general description and a brief technical discussion on the three layers is presented.
%
The first layer is represented by the \emph{trajectory generator} and it generates the desired footsteps and the desired divergent component of motion (DCM) \cite{Englsberger2011} trajectories. In the second layer, a control law based on \emph{simplified robot models} guarantee the tracking of the desired DCM, CoM and zero moment point (ZMP) trajectories. Finally, the third control layer is given by the \emph{whole-body QP} inverse kinematics. This layer ensures the tracking of the desired feet position and orientation, the desired CoM trajectories, and also the desired hands position and orientation. 

\subsubsection{Humanoid Robot Models}
Before the description of the three layers, we briefly summarize the notation and the  models used for describing and controlling the robot motions. 
\begin{itemize}
    \item $\mathcal{I}$ and $\mathcal{B}$ denotes the inertial and the robot base (e.g. the pelvis) frames;
    \item  $\prescript{\mathcal{I}}{}{p}_\mathcal{B} \in \mathbb{R}^3$ is the the position of the frame $\mathcal{B}$ w.r.t $\mathcal{I}$;
    \item  $\prescript{\mathcal{A}}{}{R}_\mathcal{B} \in SO(3)$ represents the rotation matrix between the frames $\mathcal{A}$ and $\mathcal{B}$;
    \item $\prescript{\mathcal{A}}{}{\omega}_\mathcal{B} \in \mathbb{R}^3$ is the angular velocity between frame $\mathcal{B}$ and $\mathcal{A}$, expressed in $\mathcal{A}$;
    \item the \emph{skew} operator is  $\text{sk} :\mathbb{R}^{3 \times 3} \to \mathfrak{so}(3)$, $\text{sk}(A) := (A - A^\top)/2$;
    \item $s$ and $\dot{s}$ are used to represent respectively the joints angle and velocity;
    \item the robot configuration is determined by the triplet  $q = (\prescript{\mathcal{I}}{}{p}_\mathcal{B}, \prescript{\mathcal{I}}{}{R}_\mathcal{B}, s)$;
    \item the triplet $\nu = (\prescript{\mathcal{I}}{}{\dot{p}}_\mathcal{B}, \prescript{\mathcal{I}}{}{\omega}_\mathcal{B}, \dot{s})$ represents the system velocity;
    \item the Jacobian $J_{\mathcal{A}}(q)$ is the map between the robot velocity and the linear and angular velocities of the frame $\mathcal{A}$, i.e. $\prescript{\mathcal{I}}{}{v}_\mathcal{A} = J_{\mathcal{A}} \nu$.
\end{itemize}
\par
Under the hypothesis of a CoM at constant height, the motion of the humanoid robot is approximated by means of the well known \emph{Linear inverted pendulum model} (LIPM). Which dynamic equation holds \cite{Kajita2001}:
\begin{equation}
  \label{eq:3d-lipm}
  \ddot{x} = \omega^2(x - r^{zmp}),
\end{equation}
where $x \in \mathbb{R}^2$ is the vector containing the projection of the CoM on the walking surface, $r^{zmp}\in \mathbb{R}^2$ is the position
of the ZMP and $\omega$ is the inverse of the pendulum time constant.

Analogously, one can define the divergent component of motion (DCM) \cite{Englsberger2015}:
\begin{equation}
  \label{eq:dcm_def}
  \xi = x + \frac{\dot{x}}{\omega}.
\end{equation}
Clearly, the DCM time derivative is given by:
\begin{equation}
  \label{eq:dcm_dynamics}
  \dot{\xi} = \omega(\xi - r^{zmp}).
\end{equation}

\subsubsection{Trajectory Optimization Layer}
The main purpose of this layer is to evaluate the desired feet and DCM trajectories.  
\par
To plan the desired footstep positions, the humanoid robot is approximated as a unicycle \cite{8594277}. The feet are represented by the unicycle wheels, and the footsteps can be obtained through the sampling of the unicycle trajectories. 
Once the footsteps are planned, the desired feet trajectory is obtained by cubic spline interpolation.
\par
During the single support phase (i.e. only one foot is in contact with the ground), the DCM trajectory is chosen so as to satisfy the following time evolution:
\begin{equation}
\label{eq:dcm_solution_ios}
\xi^{SS} = r^{zmp} + e ^{\omega t} (\xi_0 - r^{zmp}),
\end{equation}
where $\xi_0$ is the initial position of the DCM, $r^{zmp}$ is the position of the ZMP and $t$ has to belong to the step domain $t \in [0, \; t^{step}_i]$ where $t^{step}_i$ is the duration of the $i$-th step.
\par 
In the double support phase (i.e. both feet are in contact with the ground), the DCM trajectory is generated by using polynomial function \cite{Englsberger2014} and it satisfies the following evolution:
\begin{equation}
\xi^{DS} = a_3 t^3 + a_2 t^2 + a_1 t + a_0,
\end{equation}
where the parameters $a_i$ for $i=0:3$ have to be chosen in order to satisfy the velocity and position boundary conditions.

\subsubsection{Simplified Model Control Layer}
Using the simplified models as in \eqref{eq:dcm_def}, it can be easily shown that the CoM asymptotically converges to a constant DCM, while the DCM, (see Eq.~\eqref{eq:dcm_dynamics}) has an unstable first-order dynamics.
This stabilization problem has been tackled by designing an \emph{instantaneous} controller. The authors in \cite{8625025} proposed the following control law:
\begin{equation}
\label{eq:reactive_dcm}
r^{zmp}_{ref} {=} \xi_{ref} {-} \frac{\dot{\xi}_{ref}}{\omega} {+} K^{\xi}_{p} (\xi {-} \xi_{ref}) + K^{\xi}_{i} \int{(\xi {-} \xi_{ref}) \diff t} ,
\end{equation}
where $K^{\xi}_{p}> I_2$ and $K^{\xi}_{i} > 0_2$.
\par
Once the desired ZMP position is evaluated by the DCM controller, one has to implement a ZMP controller that guarantees the tracking of the desired signal. For this purpose the authors proposed the following control law \cite{Choi2007}:
\begin{equation}
\label{eq:zmp_controller}
\dot{x}^* = \dot{x}_{ref} - K_{zmp}(r^{zmp}_{ref} - r^{zmp}) + K_{com} (x_{ref} - x),
\end{equation}
where $K_{com} > \omega I_2$  and $0_2 < K_{zmp} < \omega I_2.$

\subsubsection{Whole-body QP Control Layer}
The main control objective for the whole-body QP control layer is to guarantee the tracking of the robot Cartesian task by using the entire robot kinematics. To do so, we use a stack of tasks formulation and differently from \cite{8625025}, we add the tracking of the hand pose as a low priority task.
In details, the tracking of the feet poses and of the CoM position is considered as high priority tasks (hard constraint), while the torso orientation along with the hand pose are considered as a low priority task (soft constraint). 
\begin{subequations}
\label{eq:velocity_control_cost}
\begin{align}
f(\nu) = \frac{1}{2} [ & (v ^ * _ {\mathcal{T}}{-}J_{\mathcal{T}} \nu)^\top
 (v ^ * _ {\mathcal{T}}{-}J_{\mathcal{T}} \nu) + \label{eq:qp_torso}\\
& (v ^ * _ {\mathcal{H}_l}{-}J_{\mathcal{H}_l} \nu)^\top
K_{\mathcal{H}_l} (v ^ * _ {\mathcal{H}_l}{-}J_{\mathcal{H}_l} \nu) + \label{eq:qp_lh}\\
& (v ^ * _ {\mathcal{H}_r}{-}J_{\mathcal{H}_r} \nu)^\top
K_{\mathcal{H}_r} (v ^ * _ {\mathcal{H}_r}{-}J_{\mathcal{H}_r} \nu) + \label{eq:qp_rh}\\
& (\dot{s} {-} \dot{s}^ * ) ^\top \Lambda (\dot{s} {-} \dot{s}^*)] \label{eq:qp_postural}. 
\end{align}
\end{subequations}
Where the term~\eqref{eq:qp_torso}, with $K_{\mathcal{T}} > 0$  and $v^*_{\mathcal{T}} = -K _{\omega  _{\mathcal{T}}} \text{sk}(\prescript{\mathcal{I}}{}{R} _{\mathcal{T}} {\prescript{\mathcal{I}}{}{R}_{\mathcal{T}} ^{*}}^{\top})^{\vee}
$, tries to stabilize torso orientation to desired orientation \cite{Olfati-Saber:2001:NCU:935467}.
\par
The terms \eqref{eq:qp_lh} and \eqref{eq:qp_rh}, with $K_{\mathcal{H}_\#} > 0$ stabilizes the hand pose to the desired pose retrieved from the virtual reality application. More specifically $v ^ * _ {\mathcal{H}_\#}$ is chosen as:
\begin{equation}
\label{handVelocitiesStar}
v ^ * _ {\mathcal{H}_\#} = 
\begin{bmatrix}
K^p _{x _{\mathcal{H}_\#}} e^p_{\mathcal{H}_\#}
+ K^i _{x _{\mathcal{H}_\#}}\int{ e^p_{\mathcal{H}_\#} \diff t}\\
K _{\omega  _{\mathcal{H}_\#}} \text{sk}(\prescript{\mathcal{I}}{}{R}_{\mathcal{H}_\#} \prescript{\mathcal{I}}{}{R} _{\mathcal{H}_\#} ^{*^\top} )^{\vee}
\end{bmatrix}.
\end{equation}
Here $e^p_{\mathcal{H}_\#} = \prescript{\mathcal{I}}{}{p}_{\mathcal{H}_\#}
- \prescript{\mathcal{I}}{}{p}^*_{\mathcal{H}_\#}$, the gains $K^p _{x _{\mathcal{H}_\#}}$, $K^i _{x _{\mathcal{H}_\#}}$, $K _{\omega _{\mathcal{H}_\#}}$ are positive definite matrices. $\prescript{\mathcal{I}}{}{p}^*_{\mathcal{H}_\#}$ and $\prescript{\mathcal{I}}{}{R} _{\mathcal{H}_\#} ^{*}$ represents , respectively, the position and the orientation of the hands frames, retrieved by the virtual reality application, and expressed with respect to the robot inertial frame.
\par
The postural task \eqref{eq:qp_postural}, with $\Lambda> 0$, is achieved by asking for a desired joints velocity that depends on the error between the desired and measured joints position 
\begin{equation}
\label{eq:regularization_term}
\dot{s}^ * = -K_{s} (s - s^d),    
\end{equation}
where $K_{s}$
is a positive definite matrix.
\par
The hard constraints are:
\begin{equation}
\label{eq:velocity_control_jacobians}
J_{\mathcal{C}}(\nu) \nu = v^ * _ {\mathcal{C}}, \quad
J_{\mathcal{F}_l}(\nu) \nu = v^ * _ {\mathcal{F}_l}, \quad
J_{\mathcal{F}_r}(\nu) \nu = v^ * _ {\mathcal{F}_r},
\end{equation}
where $v^*_{\mathcal{C}}$ is the linear velocity of the CoM, $v^*_{\mathcal{F}_l}$ and $ v^*_{\mathcal{F}_r}$ are respectively the
desired left foot and right foot velocities. More specifically $v^*_{\mathcal{F}_\#}$, where
$\#=\{l, r\}$, is chosen as:
\begin{equation}
\label{feetVelocitiesStar}
v^*_ {\mathcal{F}_\#} = \prescript{\mathcal{I}}{}{\dot{p}}^*_{\mathcal{F}_\#} -
\begin{bmatrix}
K^p _{x _{\mathcal{F}}} e^p_{\mathcal{F}_\#}
+ K^i _{x _{\mathcal{F}}}\int{ e^p_{\mathcal{F}_\#} \diff t}\\
K _{\omega  _{\mathcal{F}}} \text{sk}(\prescript{\mathcal{I}}{}{R}_{\mathcal{F}_\#} \prescript{\mathcal{I}}{}{R} _{\mathcal{F}_\#} ^{*^\top} )^{\vee}
\end{bmatrix}.
\end{equation}
Here $e^p_{\mathcal{F}_\#} = \prescript{\mathcal{I}}{}{p}_{\mathcal{F}_\#}
- \prescript{\mathcal{I}}{}{p}^*_{\mathcal{F}_\#}$, while the gains are positive definite matrices. $\prescript{\mathcal{I}}{}{p}^*_{\mathcal{F}_\#}$ and $\prescript{\mathcal{I}}{}{R} _{\mathcal{F}_\#} ^{*}$ represents , respectively, the position and the orientation of the feet frames, retrieved by the trajectory optimization layer.
\par
Finally, the desired velocity of the CoM $v^ * _{\mathcal{C}}$ is chosen as:
\begin{equation}
v^*_{\mathcal{C}} = \dot{x}^* - K^p _{\mathcal{C}}(x - x^*) - K^i _{\mathcal{C}}\int{ x - x^*  \diff t},
\end{equation}
where the gain matrices are positive definite, $\dot{x}^*$ is the output of the ZMP-CoM \eqref{eq:zmp_controller} controller and $x^*$ is the integrated signal. 

\section{Human Adaptability and Embodiment of the Robot}
One of the goals of this paper is to, rather informally,  extend the work presented in the literature, namely pertaining to measuring how humans can adapt while teleoperating the robot when the dynamics of the robot can not be ignored, consequently we also try to answer the question: \textit{how immersive, and intuitive is this teleoperation set-up?}
These aspects can be investigated thanks to the whole-body immersion experience that our architecture is designed for.
And while we do not take advantage of the \textit{human learning for robot skill synthesis} paradigm introduced in \cite{doi:10.1177/1059712311411112}, our goal is to assess the performance of the human operator, and her/his ability to produce the appropriate commands, adapt to the delayed response of the robot and mitigate the visual and feedback latency issues discussed below.

We design our experiments in order to attempt to answer our question pertaining to the level of immersion obtained using the proposed architecture, as illustrated in the following subsections. The actual experiments and results will be postponed to Sect.~\ref{sec:experiments}.

\subsection{The "Rubber Hand" Experiment}

Based on the seminal work in \cite{Botvinick1998}, in which scientists studied interaction between vision, touch, and proprioception, we perform a similar experiment, with the difference being the "rubber hand" is replaced with the robot arm, and the human is not directly looking at the arm, but rather "through the robot eyes" utilizing the VR set-up, further strengthening the immersion level. We then employ a modified, more simple \enquote{questionnaire}\footnote{We mainly focus on the participants ascertaining whether or not they agree with certain statements.} to the one reported in the paper above, the results of which are discussed in Sect.~\ref{sec:experiments}.

\subsection{Locomotion Speed}

We investigated how the human operator adapts to the robot speed. Robots, specially humanoids that are smaller in size with respect to the average human operator, tend to walk slower, and this limitation might interfere with the operators ability to command the robot effectively. A locomotion teleoperation experiment will address this issue. One expects the operator to neglect the robot speed limitations, at least in the first couple of trials.
Training is thus required to make the operator not only adapt at using the "friction-less" omni-directional mill, but also to generate commands that are appropriate for the robot, and not speed up unnecessarily.

\subsection{Visual Feedback Latency}

One key aspect pertaining to the embodiment of the robot is the visual feedback obtained through the Oculus head mounted set - HMD. Indeed, there is a difference, not just in terms of the field of view and resolution with respect to the human naked eye, but also the latency of the captured images and the ability to control the head adequately.

This aspect is interesting if one has to obtain a "smooth" teleoperation experience, and for this reason it is important to assess whether users  adapt quickly to these latency issues.

\section{Experiments and Results}
\label{sec:experiments}
In this Section, we present experiments obtained with the architecture shown in Figure~\ref{fig:block-diagram}. We use the iCub \cite{Metta2010} \cite{Natale2017}, a $\SI{1.04}{\meter}$ tall humanoid robot, to carry out the experimental activities. 
The walking control architecture runs on the on-board computer inside of the robot, namely a 4-th generation Intel Core i7 @ $\SI{1.7}{\giga \hertz}$, while the virtual reality and the omnidirectional treadmill applications run on a Windows machine equipped with a 8-th generation Intel Core i7 @ $\SI{4.1}{\giga \hertz}$.
\par
To validate the proposed architecture, we decided to perform two main experiments, which are used
as benchmarks for both the tracking capabilities and for the user embodiment. Namely:
\begin{itemize}
    \item [-] \textbf{Experiment 1}: the user walks inside the omnidirectional treadmill, while the user hands and head are mapped onto the robot hands and head respectively \footnote{The test is illustrated in the following video: \url{https://www.youtube.com/watch?v=jemGKRxdAM8}}; 
    \item [-] \textbf{Experiment 2}: the user is submitted to a \emph{rubber hand-like experiment} \cite{Botvinick1998}.
\end{itemize}


\subsection{Manipulation through Upper-body Retargeting and Locomotion}
\begin{figure}[t]
  \begin{subfigure}[b]{0.32\textwidth}
        \centering
        \includegraphics[width=\textwidth]{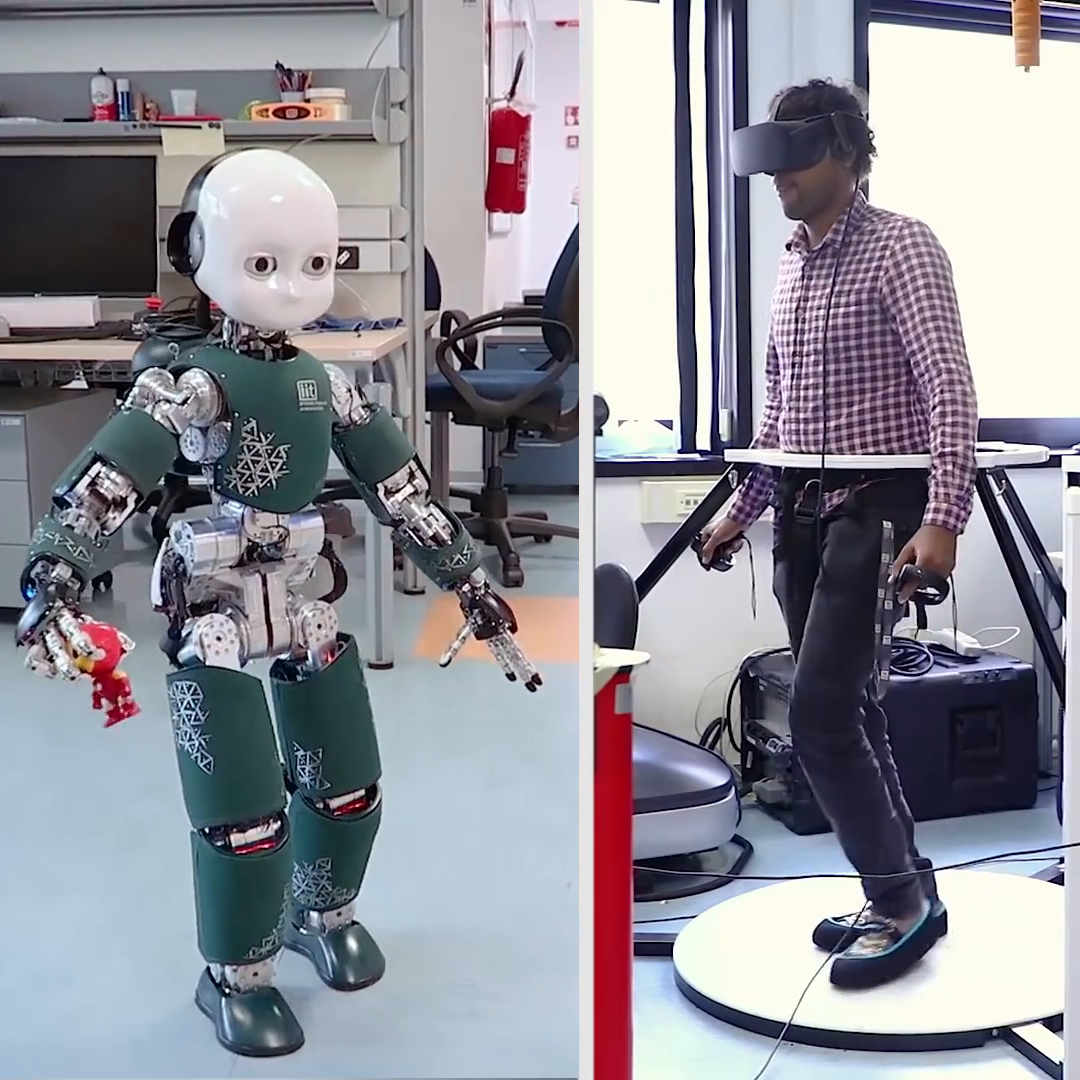}
    \end{subfigure}
    \hfill
     \begin{subfigure}[b]{0.32\textwidth}
        \centering
        \includegraphics[width=\textwidth]{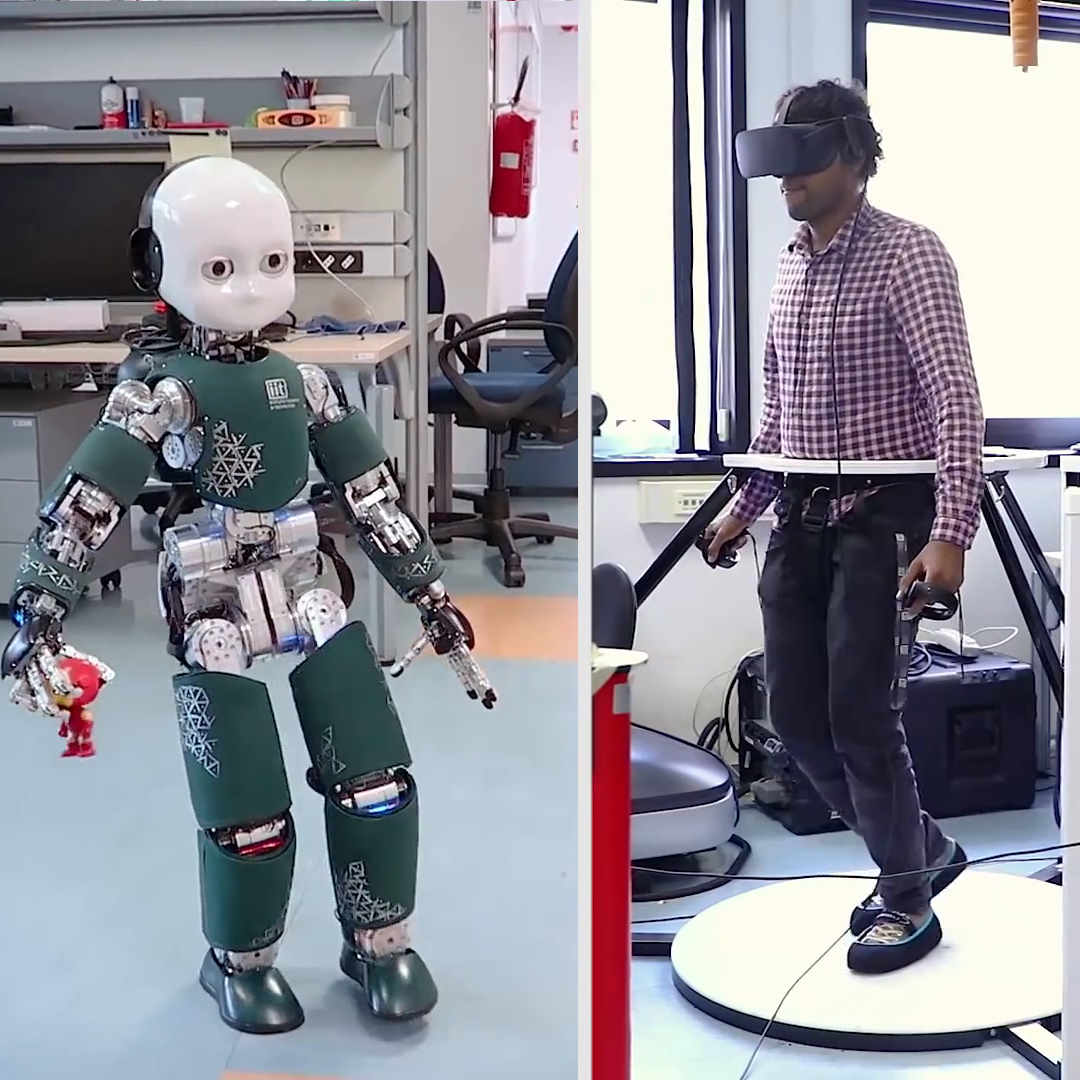}
    \end{subfigure}
    \hfill
    \begin{subfigure}[b]{0.32\textwidth}
        \centering
        \includegraphics[width=\textwidth]{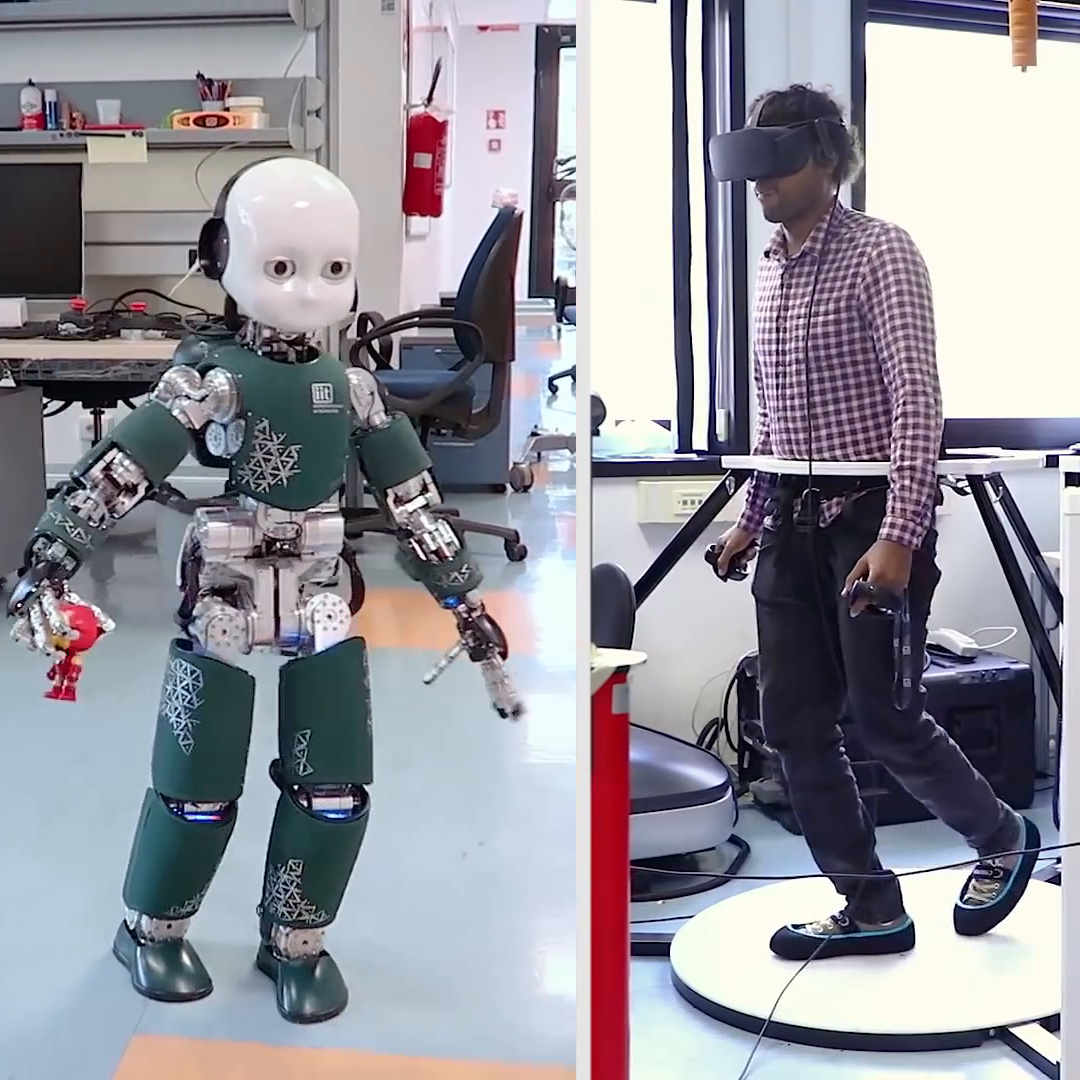}
    \end{subfigure}
    \caption{The humanoid robot is teleoperated with the architecture proposed in the paper.}
    \label{fig:icub_walking}
\end{figure}

\begin{figure}[!p]
  \begin{subfigure}[b]{0.5\textwidth}
        \centering
        \includegraphics[width=\textwidth]{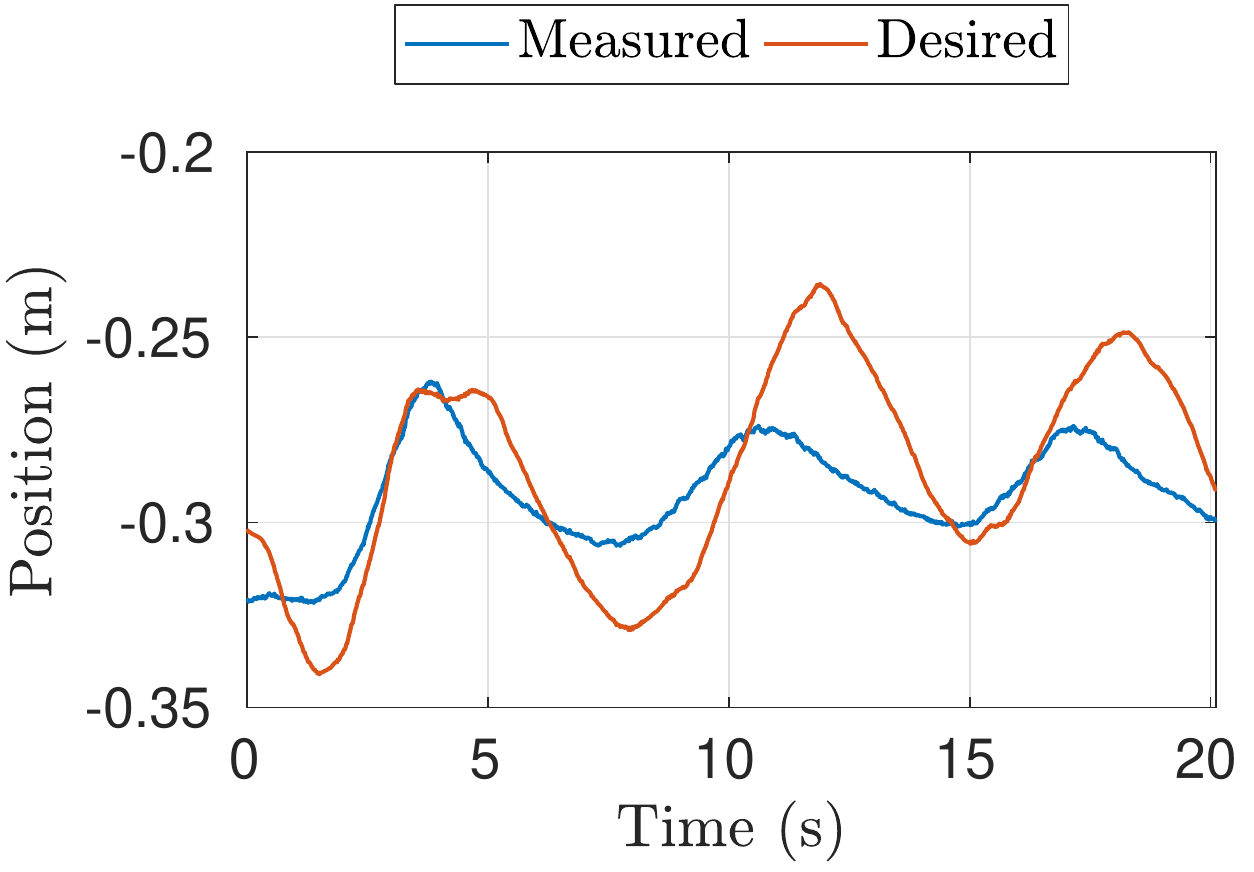}
        \caption{$x$ component.}
    \end{subfigure}
    \hfill
     \begin{subfigure}[b]{0.5\textwidth}
        \centering
        \includegraphics[width=\textwidth]{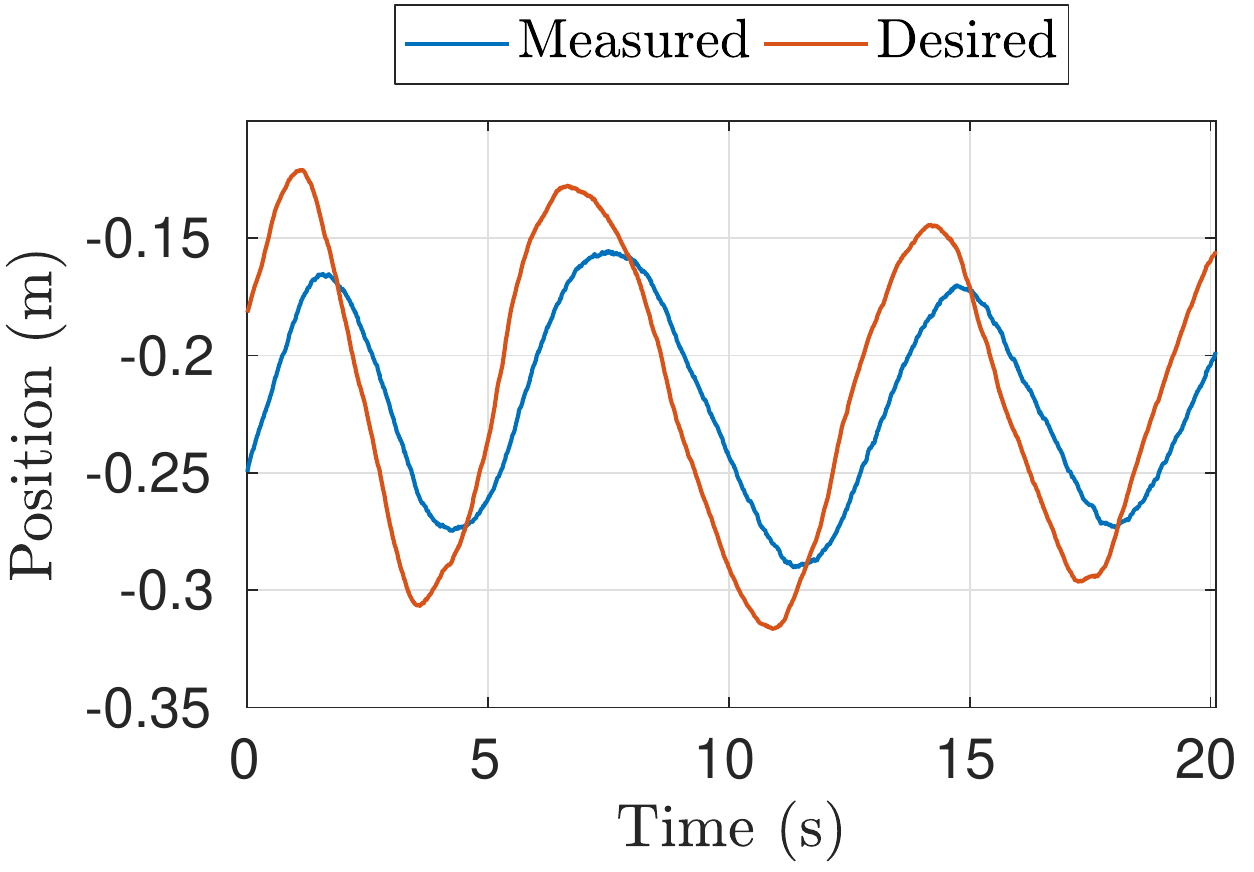}
        \caption{$y$ component.}
    \end{subfigure}\\
    \\
    
    \begin{subfigure}[b]{0.5\textwidth}
        \centering
        \includegraphics[width=\textwidth]{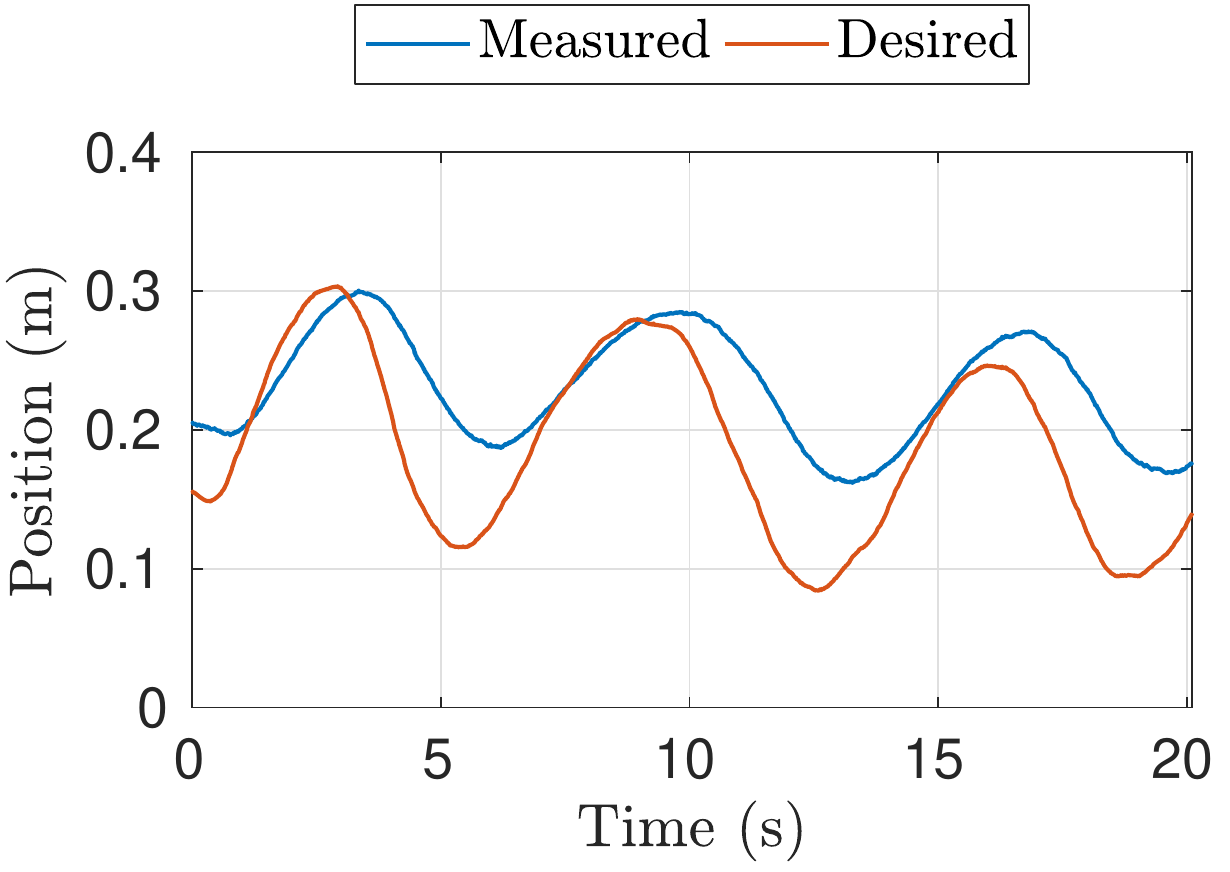}
        \caption{$z$ component.}
    \end{subfigure}
      \begin{subfigure}[b]{0.5\textwidth}
        \centering
        \includegraphics[width=\textwidth]{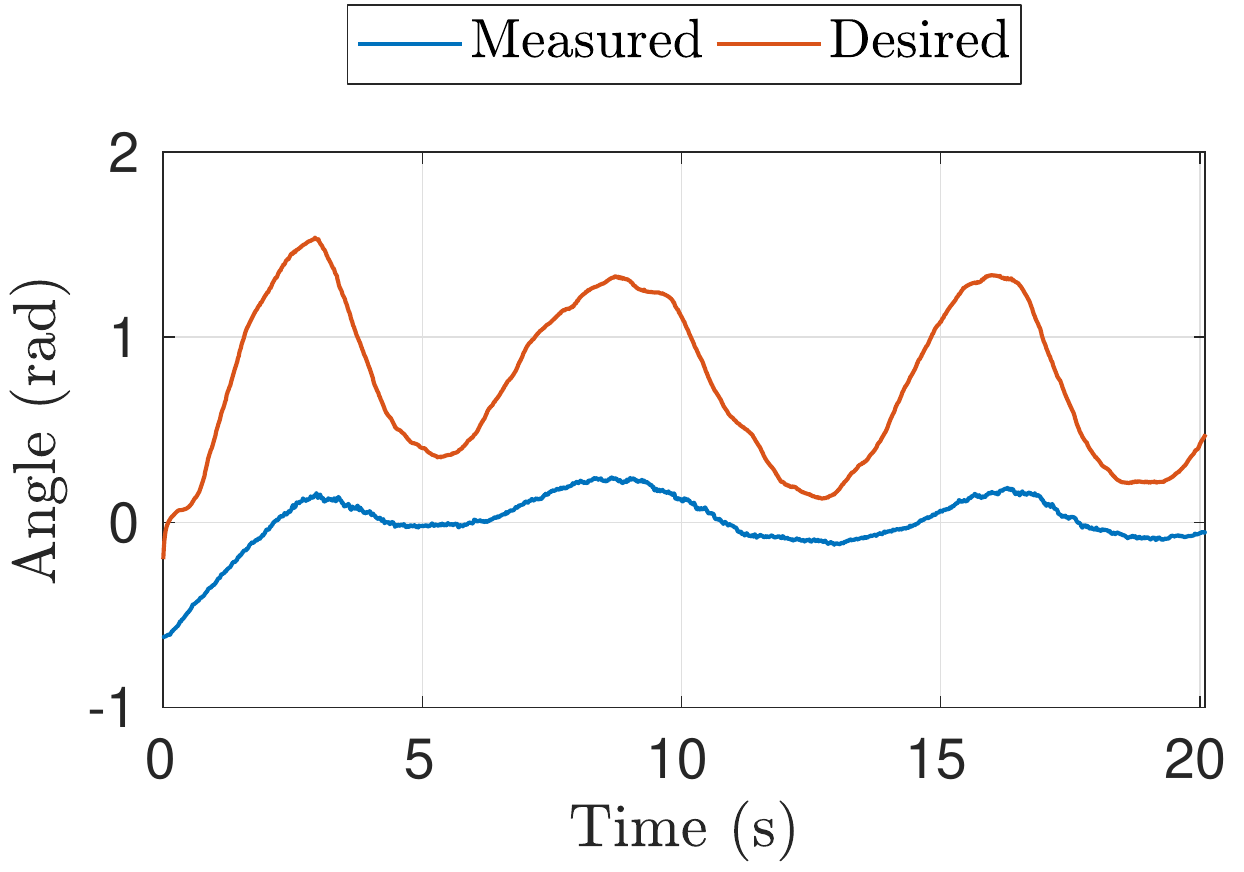}
        \caption{Roll angle.}
    \end{subfigure}\\
    \\
    
     \begin{subfigure}[b]{0.5\textwidth}
        \centering
        \includegraphics[width=\textwidth]{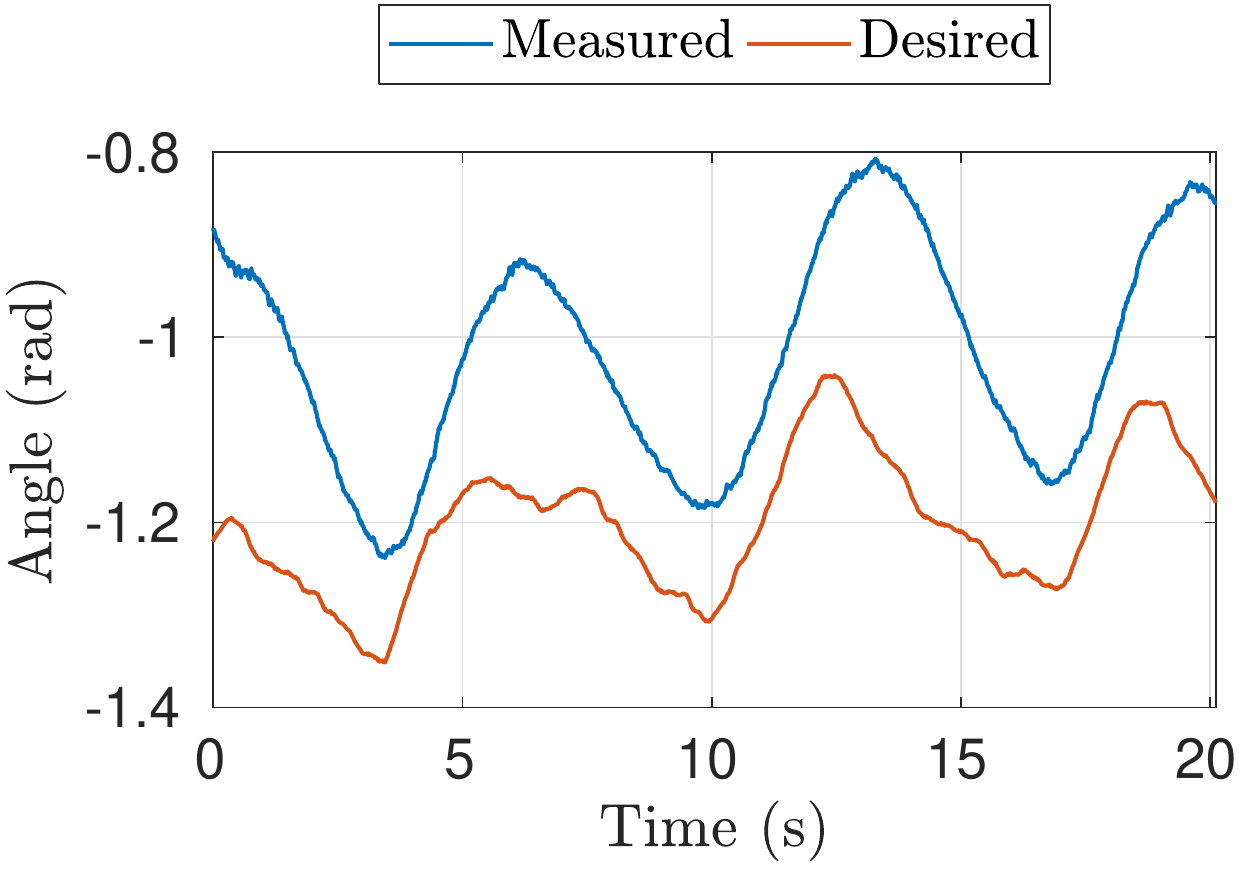}
        \caption{Pitch angle.}
    \end{subfigure} 
    \hfill
    \begin{subfigure}[b]{0.5\textwidth}
        \centering
        \includegraphics[width=\textwidth]{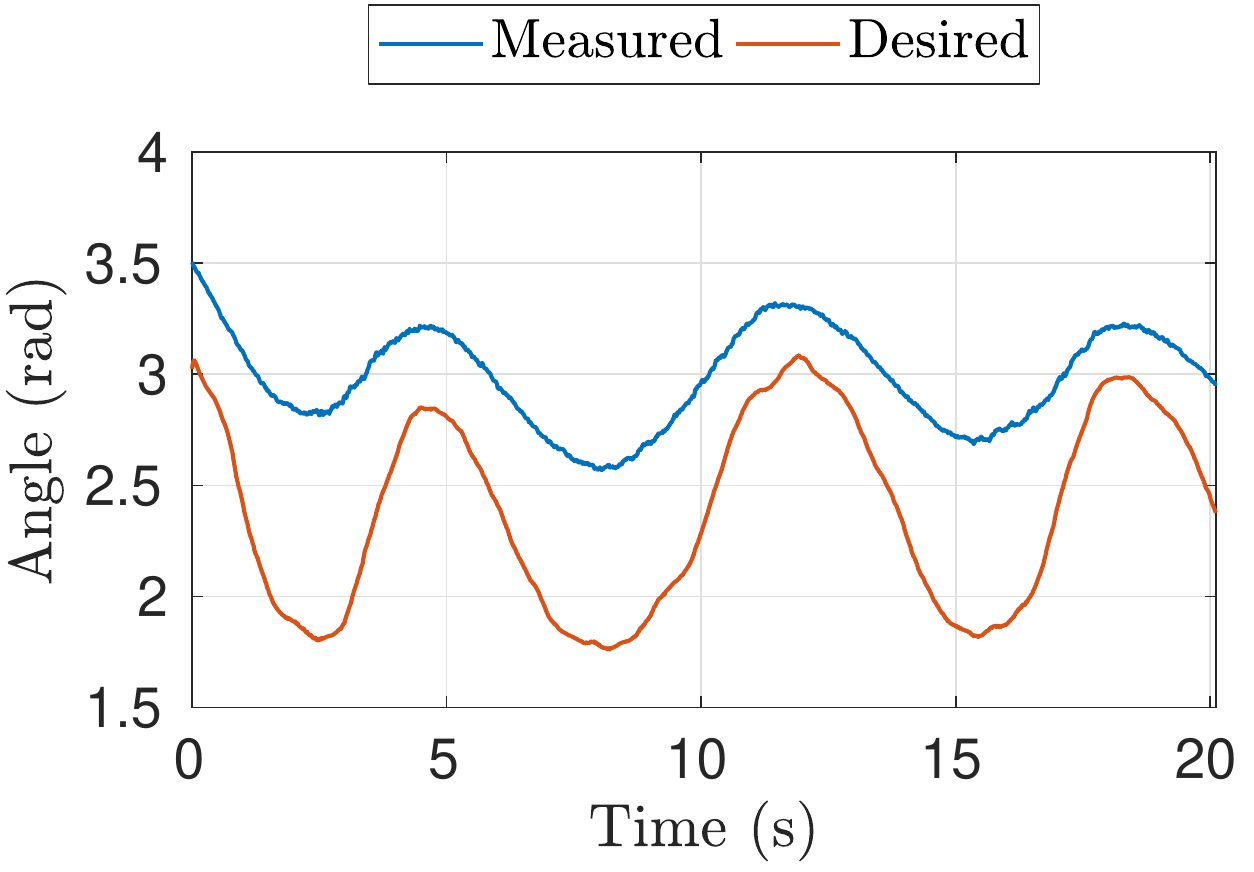}
        \caption{Yaw angle.}
    \end{subfigure}
    \caption{Tracking of the desired left hand pose.}
    \label{fig:hand_pose}
\end{figure}

In this Section, we validate the teleoperation architecture, namely the one described in Sect.~\ref{sec:teleoperation_architecture}, from the tracking point of view. The user commands the robot to walk in a domestic-like scenario (i.e. obstacle are present in the robot field of view).
During this experiment, we verified that the robot is able to walk following the directions given by the omnidirectional treadmill, however, it is also clear that the robot does not walk at the same pace as the human user, given that no footstep retargeting is performed. As a matter of fact, six test subjects have been asked to try to make the robot walk using the treadmill, while all of them showed some difficulty in keeping constant walking at first, they eventually learned how to drive the robot within a short time. With the Oculus, i.e. visual feedback from the robot, many users asserted that they felt as if they were moving in space, despite being walking on the same spot.
\par
Figure~\ref{fig:hand_pose} shows the tracking of the hand pose. The blue line represents the desired position and the Euler angles expressed with respect to the robot root link, i.e. the pelvis. The orange line is the current hand pose, i.e. the homogeneous transformation of the hand obtained from forward kinematics using joint encoder values measured from the robot. 
Notice the \emph{large} tracking error, which is mainly due to the upper body retargeting being treated as the second task as in \eqref{eq:qp_lh} and \eqref{eq:qp_rh}. 
Another explanation for the tracking error comes out from the conflict between the postural task,~\eqref{eq:qp_postural}, and the hand Cartesian task, ~\eqref{eq:qp_lh} and \eqref{eq:qp_rh}. Indeed the effort of the postural task, i.e. to keep the joints angle in a predefined constant position, could be in contrast with hand tracking term and, as a consequence, the tracking error will be a function of the weights and the gains chosen in the whole-body QP control layer. 
To mitigate this, we suggest decreasing the weight related to the postural task. Our experience, however, showed that decreasing too much this weight will bring to undesired oscillations in the arms.  

\subsection{Proprioception and the Robot Hand Feel}
Our experiment differs from the traditional \emph{rubber hand experiment} in the following ways;
\begin{itemize}
\item the robot hand is not covered in such a way as to look similar to the human operator hand in shape and skin-tone;
\item the human operator will not be looking directly at the robot hand, but rather through the robot eyes, utilizing the virtual reality system.
\end{itemize}

Yet given enough time and repetition, indeed synchronization of touching both the robot hand and the human hand and repeating the motion compels the human to associate the robot hand with their own, further supporting the findings in \cite{Botvinick1998}. However, it also begs a question; \textit{to which degree does the vision affect proprioception}. This experiment indeed verifies that the level of immersion and telexistance of the proposed architecture is satisfactory.
\begin{figure}[!b]
    \centering
    \includegraphics[width=\textwidth]{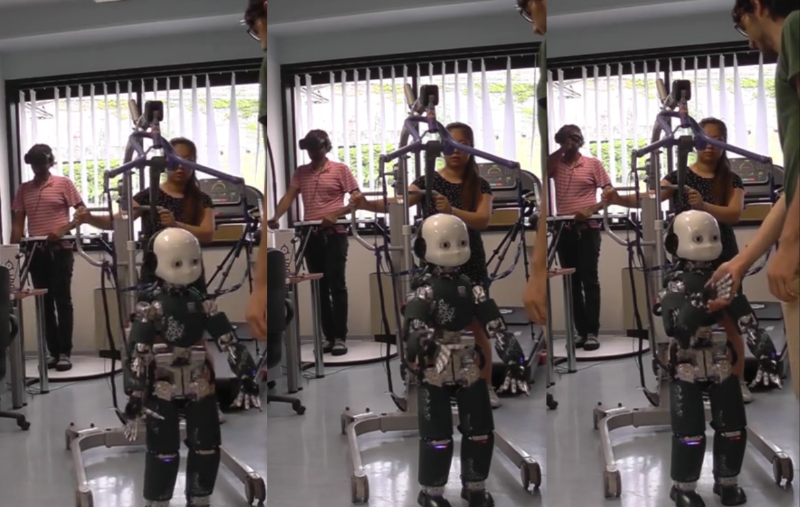}
    \caption{Right hand imitation experiment.}
    \label{fig:hand_imitation_experiment}
\end{figure}
\begin{figure}[!b]
  \begin{subfigure}[b]{0.32\textwidth}
        \centering
        \includegraphics[width=\textwidth]{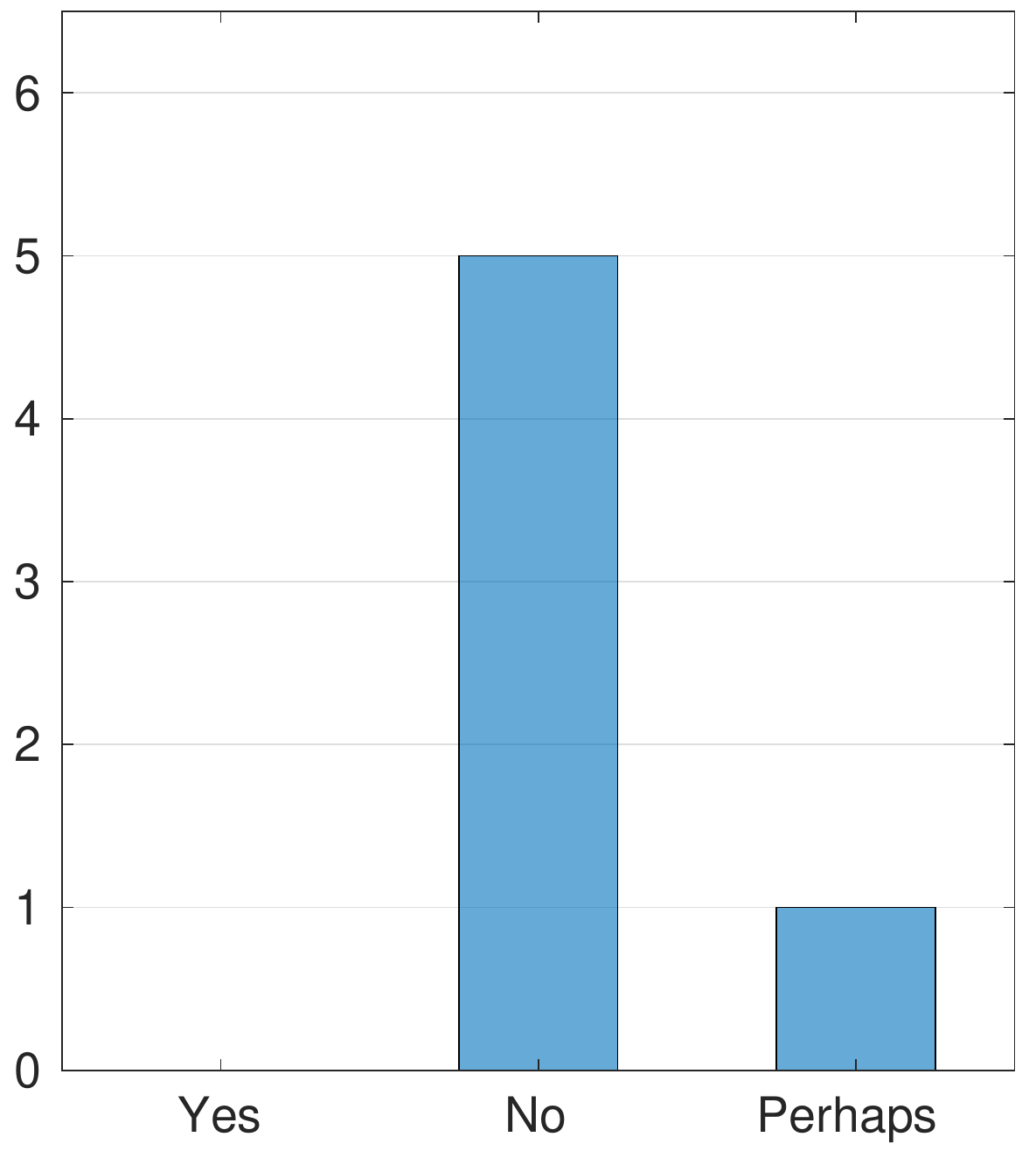}
        \caption{Statement 1}
        \label{fig:statement1}
    \end{subfigure}
    \hfill
     \begin{subfigure}[b]{0.32\textwidth}
        \centering
        \includegraphics[width=\textwidth]{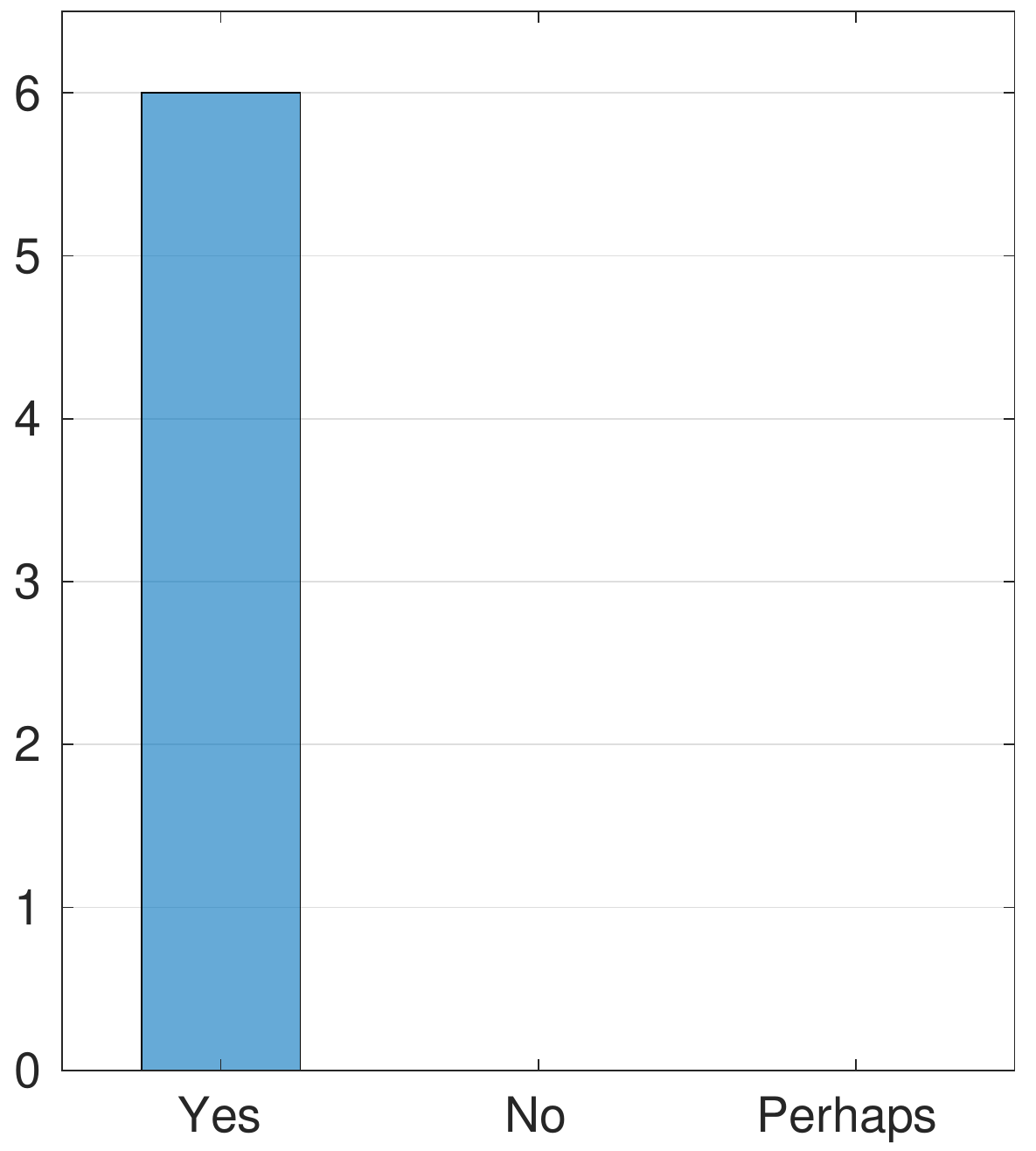}
        \caption{Statement 2}
        \label{fig:statement2}
    \end{subfigure}
    \hfill
    \begin{subfigure}[b]{0.32\textwidth}
        \centering
        \includegraphics[width=\textwidth]{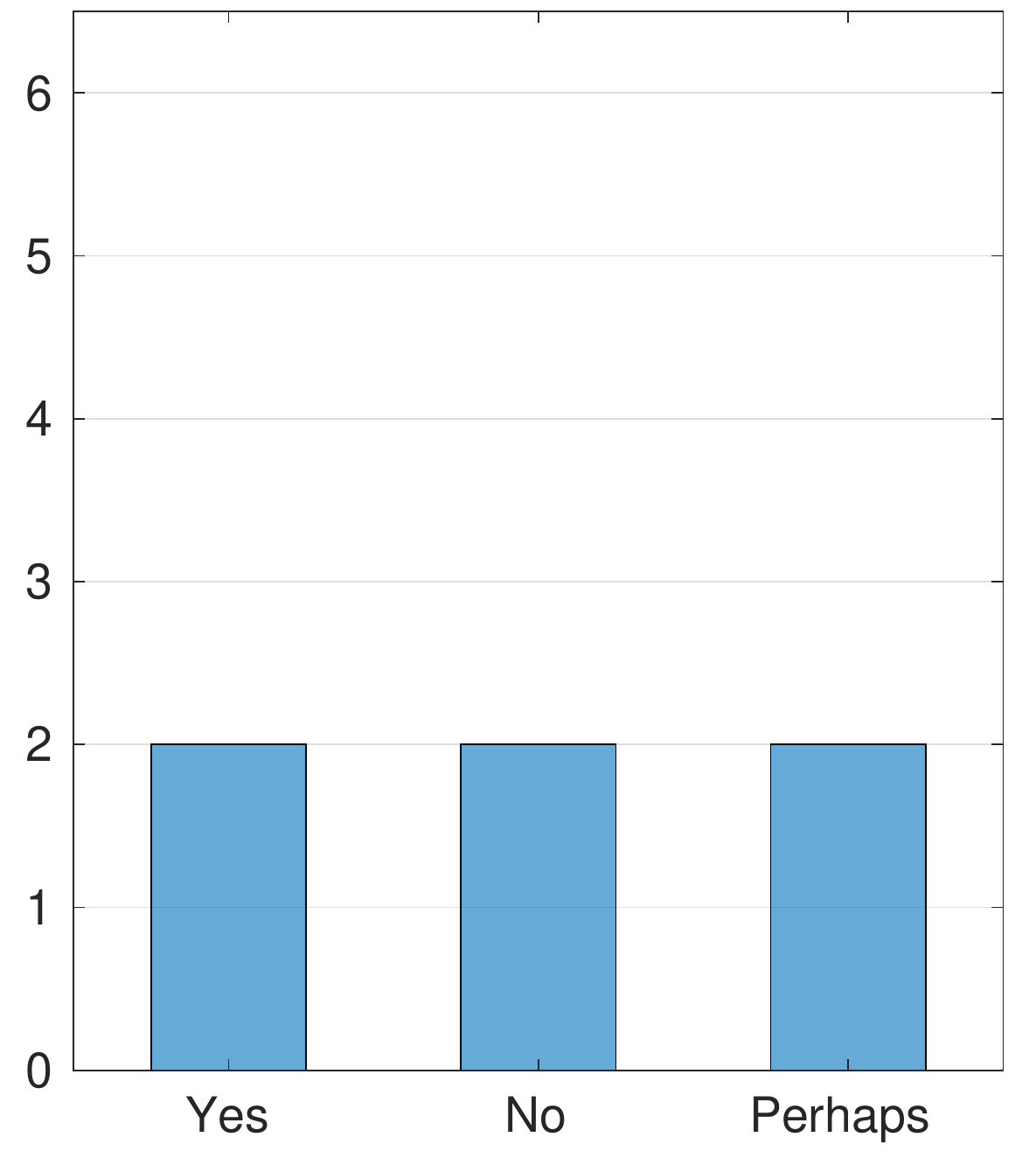}
        \caption{Statement 3}
        \label{fig:statement3}
    \end{subfigure}
    \caption{Robot hand experiment statements.}
    \label{fig:statements}
\end{figure}

\begin{figure}[thb]
  \centering
  \includegraphics[width=\textwidth]{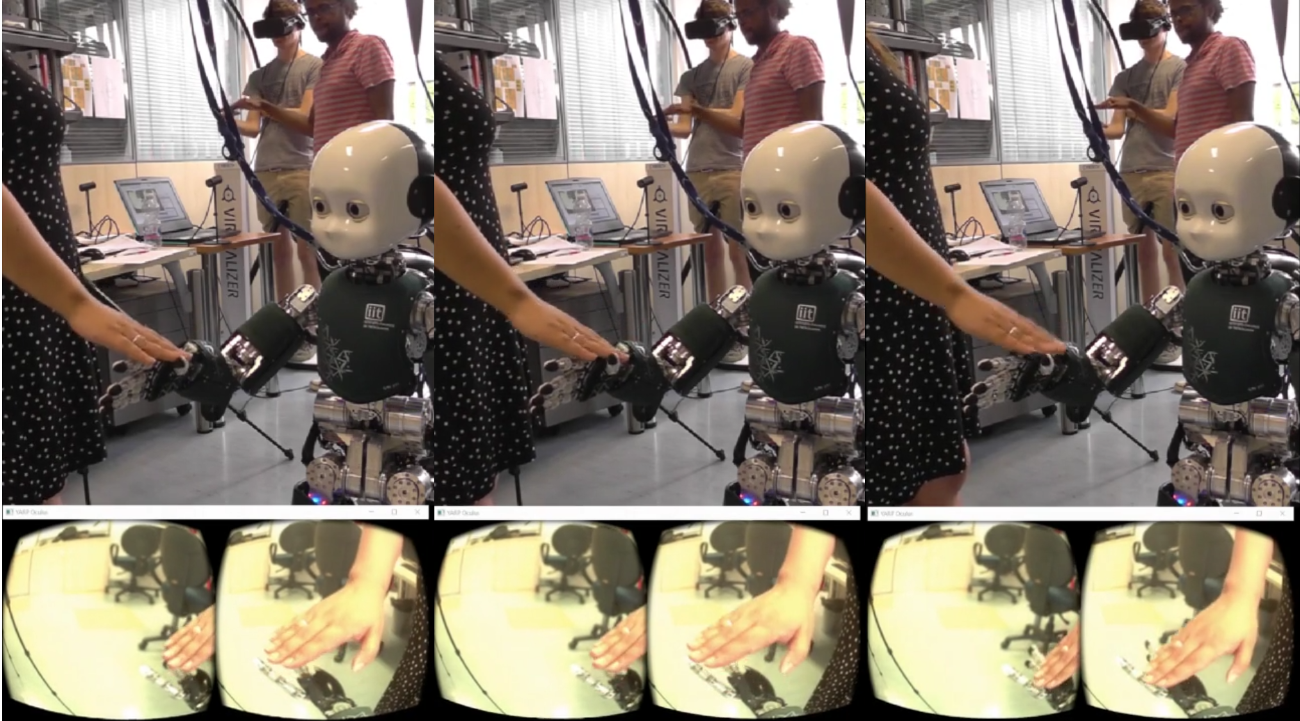}
  \caption{Right hand embodiment}
  \label{figure::RubberHand}
\end{figure}
Six participants did this experiment, and after completing it, they had to answer whether they agree with the following statements or not;
\begin{enumerate}
\item The color/shape and size of the robot hand is clearly different from my own and hence I was not able to think of it (associate it with) as my own hand.
\item After some time, I felt like a person was touching my own hand.
\item When they pushed the robot hand,  I had the impulse to react since I thought it was my own hand being pushed.
\end{enumerate}

All participants agreed with the second statement, and almost all (5) of them didn't agree with the first statement (the person who agreed said it took him some time to associate it). Figure~\ref{fig:statements} shows the response of the participants to the above statements.

Indeed, a larger number of test subjects is necessary to draw conclusions, but from these first experiments, we can observe that the users could associate the robot hand to their own hands, despite that the shape, the colors and the mechanical structure were different to that of the human. This is also the main difference between our experiment and the original rubber hand experiment. These aspects motivates incorporation of a force feedback/haptic mechanism allowing a more comprehensive immersion.

\subsection{Discussion}
\label{sec:discussion}
We discuss here several aspects pertaining to the proposed architecture and the performed experiments, namely; limitations and possible extensions in terms of whole-body retargeting involving but not limited to also feet placements retargeting. It is rather implicit that one can replace the motion capture system presented here with any equivalent one (a system that provides kinematic quantities in the Cartesian or configuration space of the human operator), and hence, such extension is not explicitly discussed.

\subsubsection{Limitations}
We discuss here limitations of the current set-up, in a high level, rather informal manner, and to this end, note that:
\begin{itemize}
    \item The robot joints limits, and mechanical design are inehrently different from that of the human, and hence the motions of the operator are not always achievable by the robot.
    \item The oculus VR head-set and its YARP implementation, as well as streaming images through the network imposes some latency, which might indeed affect not only the immersion, but also the ease of teleoperating the robot.
    \item The robot walking speed is different from that achievable by the human, which constraints the human to adapt his/her movement to match the robot, which may induce the so-called VR sickness \cite{c40}.
    \item The omni-directional platform, and while allows for a great advantage of being able to command long distance locomotion without needing to move from one's place, yet its mechanical construction, specially the ring construction, obstructs the human ability to move the hand of the robot in certain directions. Also the way the human needs to walk in the virtualizer requires some adaptation.
\end{itemize}



\subsubsection{Whole-body Online Motions Retargeting and Teleoperation}
A possible extension to our proposed architecture could be to incorporate whole-body retargeting including lower body motions. To motivate this extension consider the following; the human is teleoperating locomotion of the robot through the virtualizer omni-directional mill remotely, and the robot comes up against stairs or some physical barrier, in this scenario, if one can trigger a whole-body retargeting module somehow, by mapping the human legs end-effectors motions to those of the robot's, the robot can climb the stairs or climb over the said barrier.

To this end, we will describe the modifications to our previous structure that allows such an extension in the following passage:

\begin{itemize}
    \item A similar interface module for the Oculus VR side.
    \item For the virtualizer side, We have an additional block, namely Legs-retargeting that is event driven, meaning, uppon initialization, the module takes over from the joypad and sends the desired legs joints configurations trajectories. Initialization can be done through the oculus joy-pads buttons (or any other means).
    \item A Retargeting module. This inherits from and replaces the walking module in the case of upper-body retargeting. The idea is that if the Legs-retargeting module is invoked, this module handles retargeting and walking with dynamic balancing, otherwise it invokes a walking controller. Indeed, implementation-wise, this module will give rise to a lot of complexities that needs to be addressed (e.g. how to handle the transition between the two tasks), however, they are not within the scope of the current work.
\end{itemize}

\section{Conclusion and Future Work}

An architecture for effectively teleoperating bipedal humanoids and telexistance was proposed in this paper, and a case study of the implementation on the iCub robot was demonstrated. Interesting aspects pertaining to the user experience and immersion were discussed through a series of experiments.

In terms of ease of use and immersion, the proposed architecture is indeed, based on the gathered users' responses, very intuitive and compelling.

This result is further supported by the short training periods to effectively be able to teleoperate the robot (on average, only three trials were required) as  well  as  by the rubber hand experiment results where all the participants
confirmed what we expected in terms of embodying the robot
and associating it with one’s own self.

Therefore, we consider these preliminary results proving that our architecture is a good starting point to guarantee successful full immersion teleoperation and telexistence.

As future work, we plan to improve the compliance of the robot in case of unexpected interaction with the environment. This will allow having a platform that can operate alongside humans to carry out collaborative tasks (e.g. \cite{Romano2018}).
Another interesting future work is to increase the level of the embodiment by developing a whole-body retargeting architecture.

\section*{Disclaimer}
The content of this publication is the sole responsibility of the authors. The European Commission or its services cannot be held responsible for any use that may be made of the information it contains.

\section*{Acknowledgment}
This project has received funding from the European Union’s Horizon 2020 research and innovation program under grant agreement No. 731540 (An.Dy).
\par
The authors would like to extend their thanks to Marco Randazzo, Andrea Ruzzenenti, Daniele Domenichelli and Lorenzo Natale for the integration of the VR Headset driver in YARP.

\bibliography{bibliography.bib} 
\bibliographystyle{bibtex/splncs03_unsrt}

\end{document}